# Evaluation of LLMs for Process Model Analysis and Optimization

Akhil Kumar[1]    J. Leon Zhao[2]    Om Dobariya[3]

**Abstract**. In this paper, we report our experience with several LLMs for their ability to understand a process model in an interactive, conversational style, find syntactical and logical errors in it, and reason with it in depth through a natural language (NL) interface. Our findings show that a vanilla, untrained LLM like ChatGPT (model o3) in a zero-shot setting is effective in understanding BPMN process models from images and answering queries about them intelligently at syntactic, logic, and semantic levels of depth. Further, different LLMs vary in performance in terms of their accuracy and effectiveness. Nevertheless, our empirical analysis shows that LLMs can play a valuable role as assistants for business process designers and users. We also study the LLM's "thought process" and ability to perform deeper reasoning in the context of process analysis and optimization. We find that the LLMs seem to exhibit anthropomorphic properties.

## 1    Introduction

Business process management plays an important role in organizations [5] because all work is performed through processes. Processes are required to handle customer orders, purchase requisitions, insurance claims, hiring/firing an employee, approving leave requests, reimbursing travel expenses, etc. Formal process design helps everyone in an organization to understand the various steps in a process. Process designers in an organization are responsible for designing complete, correct, and error-free processes in languages like BPMN [3] based on their understanding of the process acquired from domain experts. These process designs or models evolve over time as new ways are developed for executing the process. These improvements that often take advantage of technology lead to faster processing times and better customer service.

   A **Large Language Model** (LLM) is a type of artificial intelligence (AI) program that ingests large amounts of data to perform natural language processing (NLP) tasks using deep learning techniques. It is based on the transformer architecture and the attention mechanism [18]. LLMs are advanced neural networks (NNs), computing systems inspired by the human brain. They work in a question-answer mode by responding to user queries or prompts. They are also called **Generative Pre-trained Transformers** (GPTs) because they are trained on huge amounts of data available on the internet and generate responses to prompts based on the patterns in the prompts.

---

[1] Pennsylvania State University, University Park {**akhilkumar@psu.edu**}
[2] Chinese University of Hong Kong, Shenzhen {**leonzhao@cuhk.edu.cn**}
[3] Pennsylvania State University, University Park {**okd5069@psu.edu**}

In this paper, we evaluate the ability of LLM technology to understand a process model, find syntactical and logical errors in it, and reason with it in depth through a conversational interface. This new technology can open the possibility for non-expert users to interact with an LLM and check their process models for correctness, ask the LLM to make corrections to it, and perform various types of analyses on the model by themselves. This paper adopts a Design Science Research (DSR) framework that evaluates the LLM artifact for utility, consistency, and novelty.

The organization of our paper is as follows. We provide background and discuss related work in Section 2 along with a framework and a syntax for the structure of prompts. In Section 3, we undertake extensive evaluations of the ChatGPT LLM [4] on a process model from the finance domain. To validate the findings on ChatGPT, we further test the mortgage application processing case on Claude Opus 4, Grok 3, and Gemini 2.5 Flash LLMs and discuss the results. In Section 5, we report results from a healthcare process and offer a discussion on thought/reasoning processes of ChatGPT. Finally, Section 6 outlines future research directions and concludes the paper.

## 2     Background and approach

**Related work.** For several years, there has been an interest in extracting textual descriptions from process models and vice-versa [6,11,16]. The work in [13,14] developed approaches to generate textual descriptions from process models using deep learning techniques. In recent years, there has been a flurry of interest in using large language models (LLMs) to understand process descriptions and querying them to assess their capabilities.

In this new line of work [19], an initial effort was made to extract process entities and relations from textual descriptions using the GPT-3 model [1]. The authors reported that GPT-3 performed quite well at listing the activities in a process model and their precedence relationships. The model could also answer queries about the participants that performed tasks in a process model. The authors further noted that the exact text of the queries or prompts affected the quality of the answers produced by GPT-3. Subsequently, the authors developed a corpus of business process descriptions annotated with activities, gateways, actors, and flow information. A later work [2] examined the role of prompt engineering, i.e., crafting the right prompts, as a means for extracting information about a process from an LLM. They also discussed the importance of fine-tuning an LLM to prepare it for answering BPM-related queries or prompts. Another paper evaluated the suitability of LLMs for business process tasks [7], such as process mining imperative and declarative models from textual descriptions. They found that the results compared favorably with existing solutions. A comprehensive evaluation of 16 state-of-the-art LLMs from major AI vendors highlights significant performance variations across LLMs and finds a positive correlation between efficient error handling and the quality of generated models [10].

Our work is also inspired by [8,9]. In [8], it was found that 60% of users at all levels of expertise prefer LLM-generated process models to human-created ones. In [9], the authors use LLMs to redesign process models in a conversational style and report promising results. Our work goes further by investigating the LLM's ability to understand a process model in an interactive, conversational style and help non-expert users gain a better understanding of their process models.

**Framework**. A process description consists of a control flow that describes the main structures in the process, such as start node, end node, task nodes, XOR split/joins, AND split/joins, etc. It also includes task durations (temporal aspect) ranges. Moreover, there is a resource aspect that represents the role (human or machine) that performs a task. Further, the data aspect relates to input and output data of each task. A process may be subject to constraints that impose restrictions on it [5].

Table 1 shows an overall framework for this research consisting of five components: Research Questions, Dimensions Evaluated, Independent Variables, Dependent Variables, and Evaluation Method, as shown. This framework can be applied to evaluate processes from domains like business, healthcare, etc. Here, our focus is on the control flow and temporal aspects of a process. An evaluation workflow is shown in Figure 1. A process model is offered to the LLM. To assess its capabilities, a user/analyst poses prompts in an interactive, conversational style. Based on the responses, the user may ask the LLM to perform tasks such as apply the fixes to the errors, etc.

Table 1. Framework components for evaluating LLM conversational models

| Component | Description |
|---|---|
| Research Question(s) | What is the effectiveness of an LLM in analyzing, validating, and optimizing business processes in a given domain? |
| Dimensions Evaluated | Syntax validation, semantic understanding, logic error detection, reasoning ability, optimization suggestion quality |
| Independent Variables | Process complexity, prompt type/syntax, diagram notation type, domain (e.g., healthcare, mortgage) |
| Dependent Variables | Accuracy of detection, correctness of fix, completeness of response, user confidence, time to solution |
| Evaluation Method | Manual benchmarking, user testing, or alignment with expert-created gold standards |

**Interface Syntax**. The syntax of the interface is as follows.

- Describe the process in the diagram.
- Check the diagram for any logical or syntactical errors.
- Correct [Errors] and redraw the diagram.
- [Find|Calculate] the [Min|Max|Avg] Finish Time of the process.
- [Find|Calculate] [Min|Max|Avg] Finish Time by the [path_name] [path|branch].
- Consider redesign scenarios for this process… [description of possible changes]
   … Make a table describing the time and cost impact of these scenarios.
- Explain your calculation of the [Min|Max|Avg] time for [scenario #] above.
- List the inter-task constraints.
- Calculate the [Max|Min|avg] time by the surgical path if [constraint #] is relaxed.

This syntax is typical. While LLMs are known to be sensitive to exact prompt wording [15], minor variations in wording do not affect the results. Next, we describe a process in BPMN notation [3] and discuss how ChatGPT model o3 was used to analyze it in depth.

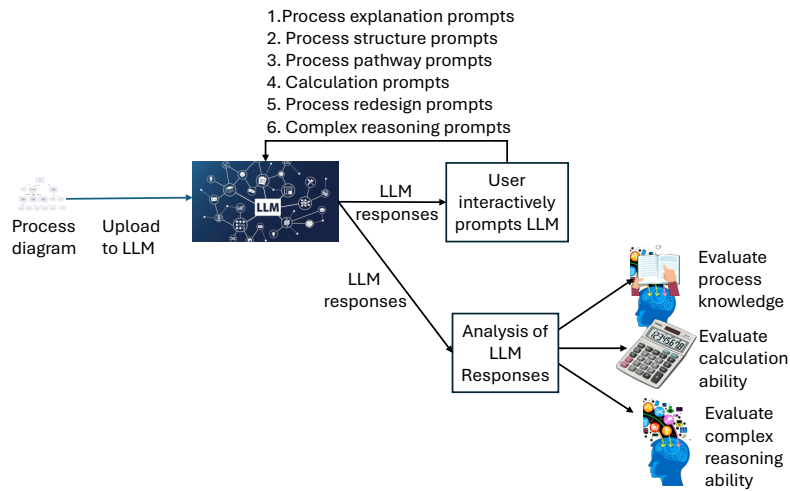

Figure 1. An architecture/workflow to evaluate LLMs for process knowledge

## 3    A mortgage application review process

In this section, we shall evaluate ChatGPT (o3) model o3 from Open AI [4] using the example mortgage process of Figure 2. We chose model o3 because it has strong analysis and reasoning capabilities. It also goes through a "thought" process and shares its "thinking" in the output it generates.  No prior training was given in this zero-shot setting.

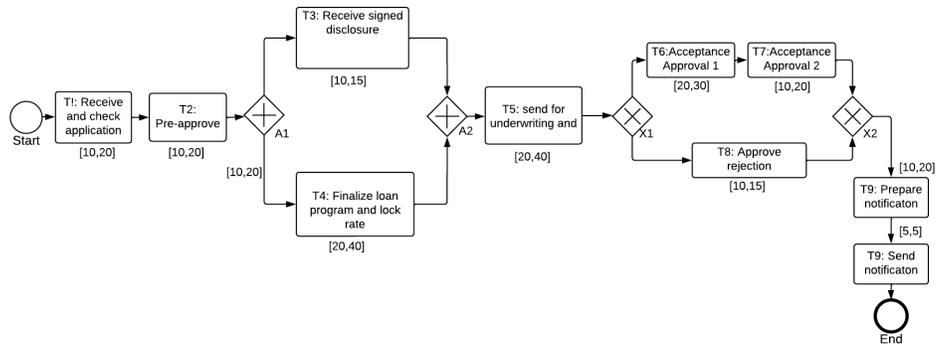

**Figure 2.** A mortgage application approval process (with minor intentional errors)

In the mortgage application process of Figure 2, after an application is received, it is checked (T1) and the applicant is then pre-qualified (T2). Subsequently, in parallel: the signed disclosures and documents are received from the client (T3); and a loan program is finalized along with the rate (T4). Next, the application is sent for underwriting (T5). If the application is acceptable to the underwriter (the upper path at the XOR node), then two further approvals are required by the loan officer (T6) and the VP (T7). However, if the underwriter declines, then the loan officer rejects the application (T8). Finally, in both cases, the relationship manager prepares a notification (T9), and it is sent to the applicant (T10). The pairs of diamonds with the '+' symbols denote parallel branches that split and merge, while the diamonds with the 'X' symbols denote choice or alternate paths in BPMN notation. The figure also shows the duration of each task as a [min, max] range.

To test basic process knowledge of o3, the process model of Figure 2 was copied and pasted into the prompt box along with the text: "Describe the process in the diagram."

**o3 Response**: The diagram depicts a **business process model** for a **loan approval workflow**, using BPMN-style notation. Here's a step-by-step description:
- **Main Process Flow …**
- **Approval Path…**
- **Rejection Path…**

- **Notification…**

Descriptions are provided for each of these sections by o3, and a summary table of tasks is given afterwards showing each task and its duration range. The output was correct. Next:

**Prompt**: Check the diagram for any logical or syntactical errors.

In its response, o3 generated the output shown in Table 2. Here it has noted the errors in our model and suggested how to fix them. It also generated three "potential modelling/logic concerns", e.g. that in many lending policies the disclosure (T3) must be signed **before** the rate can be locked (T4); and that T9 and T10 could be combined. However, these are very specific to modeling in the context of the policies of the organization.

Table 2. "Syntactic (BPMN-notation) problems" generated by o3

| # | Issue | Why it is wrong | Suggested fix |
|---|---|---|---|
| S-1 | **Duplicate task ID "T9"** is used twice ("Prepare notification" and "Send notification"). | Every element in a BPMN model must have a unique identifier. | Rename, e.g. "T9: Prepare notification" → **T9**, "T9: Send notification" → **T10**. |
| S-2 | **Gateway A1 carries a duration label [10, 20]**. | Gateways are routing objects; they never have execution time. | Delete the duration text (or, if it belongs to T2, move it back under T2). |
| S-3 | Misspelling–"notificaton" in the second T9 label. | Typographical error. | Correct to *notification*. |
| S-4 | **Truncated task name** – "T5: send for underwriting and". | Task names should be self-contained action phrases. | Finish the caption, e.g. "T5: Send for underwriting and credit decision". |
| S-5 | Minor style: "T6:Acceptance Approval 1", "T7:Acceptance Approval 2" lack a space after colon. | Minor style: "T6: Acceptance Approval 1", "T7: Acceptance Approval 2" lack a space after the colon. | Add the space ("T6: Acceptance …", "T7: Acceptance …"). |

The next prompt and response pair is as follows:

**Prompt**: Correct S1-S5 and redraw the diagram.

**o3 Response**: I corrected items S-1 through S-5 and rebuilt the BPMN diagram. [See Table 3.]

Table 3. "What changed" summary generated by o3

| Fix | Change made |
|---|---|
| S-1 | Gave the dispatch activity its own unique ID T10 |
| S-2 | Removed the stray duration label from gateway A1 |
| S-3 | Spelling corrected to "notification" |
| S-4 | Completed the caption for T5 ("…credit decision") |
| S-5 | Added spaces after the colons in T6 and T7 |

In addition to Table 3 showing "What changed", o3 also produced a diagram with the changes for us to download (omitted here). Next, we prompted o3 to "use the corrected process now". In response, o3 produced a summary of the new process and included a "Quick logical-consistency check (post-fix)" that checked off these items: *All elements uniquely named*; *Gateways have no durations printed*; *No orphaned flows / dead-ends*; *Parallel split + join properly paired (A1 ↔ A2)*; *Exclusive decision & merge (X1 ↔ X2)*; and *Spelling / label typos*.

Next, o3 was prompted to "Find the *minimum* finish time of the process," and *maximum* and *average* in the same manner. It produced the correct values (85, 195, 122.5) and gave its reasoning along with a table and a separate explanation for the calculation in the parallel section of the process. For calculating average time, it used the mid-point values in the duration range for each task.

**Process redesign with o3**. Next, redesign scenarios were presented to o3 as follows.

**Prompt**. Consider redesign scenarios for this process.
1. Task T2 is optional and can be skipped for a cost of $50.
2. T2 can be replaced with task R2, which takes 5 units of time less than T2 but incurs a cost of $25.
3. Task T7 is optional and can be skipped for a cost of $75.
4. Task T6 is optional, and can be skipped for a cost of $50.
5. Tasks T6 and T7 can be done in parallel to save time but it incurs a cost of $10.
6. Task T7 can be merged into T6 to create a new task T67 increasing the duration of T6 by 10 units of time and increasing cost by $25.
Make a table describing the time and cost impact of these scenarios.

The o3 response showed a table with columns for row #, redesign action, extra cost, time difference (versus baseline time) and new cycle-time range. The range column gave the correct new range values for cycle time for each scenario. When prompted to find the minimum time solution and cost, o3 performed a calculation for "Fastest-possible redesign" and returned a minimum time of 75 units and cost of $50 (for the rejection path). When prompted specifically to recalculate for the acceptance path, o3 produced a minimum time of 65 units and a cost of $175. This is correct too. When asked, "Why did you give me a minimum time of 75 first?", it replied:

**o3 Response**: I gave 75 minutes first because I was answering the earlier question in the most literal way: "What combination of scenarios will lead to the minimum total time for the process?" Total (or absolute) minimum means "shortest-possible completion time for any case that can run through the model." That shortest case is the rejection path, not the acceptance path, because in the baseline model it already ends sooner (85 min vs. 105 min). [Note. This is a reasonable reply.]

**Logical design error detection.** Next, we tested how well o3 could capture process design errors. Errors were introduced in Figure 2 into the approve/reject part of the process as shown in Figure 3. In each case, o3 was prompted to "Check the diagram for logical and syntactical errors." For Figure 3(a) it noted that the incorrect parallel ("+") gateway and recommended: *Replace X1 with a data-based exclusive gateway ("X"). Attach conditions such as approved? = yes / no to the two outgoing sequence flows*. For Figure 3 (b), it correctly recognized the deadlock and recommended: *Replace X2 with an exclusive merge ("X") or remove the diamond altogether (implicit merge)*. For Figure 3 (c), o3 noted: *Either add the "X" marker inside the diamond or delete the diamond and connect both flows directly into T9 (implicit exclusive merge)*.

In summary, o3 reveals a deep understanding of this process. It could find errors including minute nuances like typos and suggest improvements based on its overall understanding of the mortgage process. Further, it knows the semantics as shown in the case of Figure 3(a) where the process is syntactically correct, yet it could detect that the two branches are alternatives to each other.

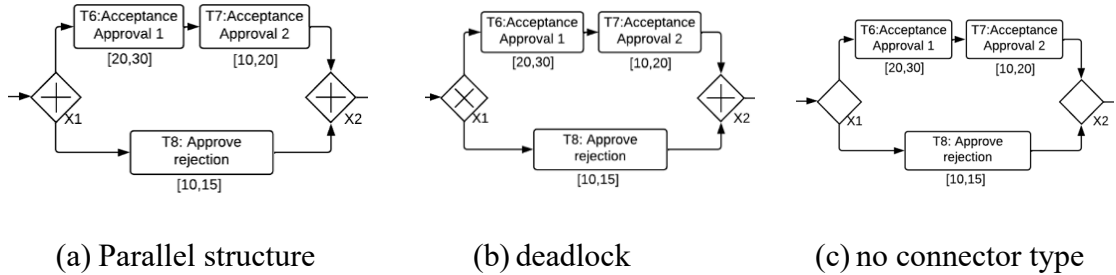

(a) Parallel structure    (b) deadlock    (c) no connector type

Figure 3. Three erroneous variant scenarios created from Figure 3

## 4 Comparative evaluation

To generalize our results on ChatGPT, we tested three other LLMs including Claude Opus 4, Grok 3, and Gemini 2.5 Flash. We used criteria such as syntax error detection, logical error detection, semantic comprehension, and reasoning ability. We compared the LLMs on these criteria using a scoring matrix as follows:

3: Fully correct and insightful
2: Mostly correct with minor omissions
1: Limited insight; superficial or partially incorrect
0: Incorrect or absent

The results are summarized in Table 4. o3 could identify all the syntax errors, while the other three LLMs failed to do so; instead, they even found non-existent errors. On logical errors, o3 and Gemini performed perfectly while the other two failed entirely. On the semantic depth criterion, o3 could calculate all the metrics – minimum, maximum, and average finish times– correctly at the very first attempt. Moreover, it could understand the process and explain its reasoning clearly. Gemini made an error in calculating the maximum time; Claude showed poor understanding of BPMN notation, while Grok gave all wrong answers. All LLMs performed perfectly on the optimization (after the correct process had been explained to them) and gave the correct answers.

When asked to reproduce the correct diagram without syntax errors, o3 did perfectly well, while all others either gave an incorrect diagram or admitted their inability directly. o3 is the best here.

Table 4. Comparison of ChatGPT with three other LLMs for the mortgage process

| Model | Syntax Error Detection | Logical Error Detection | Semantic Depth & Reasoning | Optimization Quality | BPMN Diagramming | Total Score (Max 15) |
|---|---|---|---|---|---|---|
| ChatGPT o3 | 3 | 3 | 3 | 3 | 3 | 15 |
| Claude Opus4 | 1 | 0 | 1 | 3 | 0 | 5 |
| Grok 3 | 1 | 0 | 0 | 3 | 0 | 4 |
| Gemini 2.5 Flash | 1 | 3 | 2 | 3 | 0 | 9 |

## 5  Discussion

To stress-test our approach further, we offered the medical process of Figure 4 to o3. This process describes the diagnosis of a possible femoral fracture and there are different paths based on whether the suspicion of such a fracture is valid or not after an initial exam. If so, further testing is done, and surgical or non-surgical paths (Therapy B) are pursued; otherwise, Therapy A is administered. Thus, there are nested control structures making this process more complex than the previous one. This diagram was posted in the prompt window along with the following prompt:

**Prompt**: In this BPMN process diagram the duration ranges of each task are shown next to the task. In addition, there are 5 inter-task temporal constraints, TI1, TI2, TI3, TI4, and TI5. These constraints apply to the start (end) times of tasks depending upon whether the end point of the constraint lies on the left (right) boundary of the task in the diagram. Describe this process.

In response, o3 provided a high-level narrative of the process that was correct and listed the five inter-task temporal constraints (TI1-TI5) in a table. There was a minor error (likely from image scan misinterpretation) that showed TI1 as: *T1 end → T2 end.* When this was notified to o3, it showed the corrected table.

Note that o3 could accurately check this process for logical and syntactical errors and demonstrate a deep understanding of this process. It also made valid observations about the semantics of task execution related to wait times between certain tasks like T10 and T11 where a wait is necessary. When asked to "calculate the minimum, maximum and average finish times for this process," o3 gave us Table 5. But when prompted further to calculate the Path-wise finish-time results again allowing for *maximum wait times* as allowed by the inter-task constraints, the maximum times increased by 10 because there is a possible waiting time of 10 between T1 and T2 that o3 did not consider earlier. Thus, the revised calculations reflect this as shown in parentheses in Table 5.

Again, o3 demonstrates a remarkable capability to understand a complex process with 14 tasks, nested control flow gateways and five inter-task constraints. It comprehends user questions like a human would, appreciates the nuances of process modeling and responds intelligently.

We examined its thought process from the thought text that o3 generated while answering various queries. When prompted for the minimum total time for the redesign scenarios and its cost, o3 could dissect the user prompt in considerable detail from various angles like a human would to decipher what exactly the intention of the user was. Tests on other prompts also showed that the reasoning processes seem to mimic those of humans. Even when the LLM produces a seemingly incorrect output it arises from certain underlying assumptions that are based on its mental model.

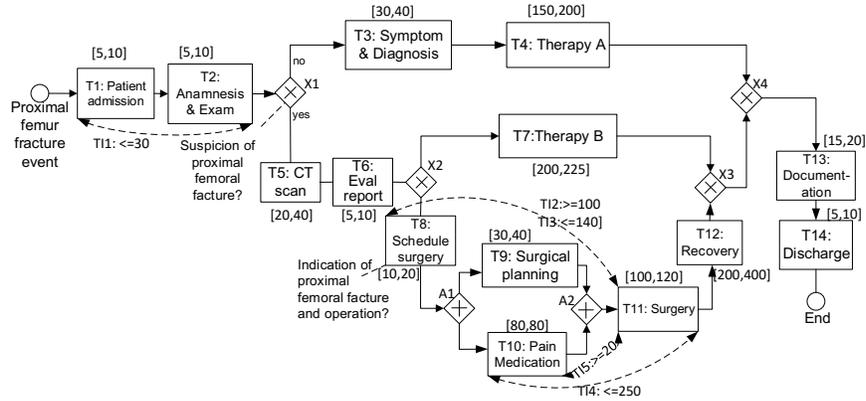

Figure 4. A healthcare process for a suspected proximal femur fracture

Table 5. Path-wise finish-time results generated by o3

| Path | Activities executed | Minimum | Maximum | Average |
|---|---|---|---|---|
| A "Conservative — no fracture"*(X1 = no)* | T1 ► T2 ► T3 ► T4 ► T13 ► T14 | 210 | 290 (300) | 250 (255) |
| B "Conservative — fracture, no surgery"*(X1 = yes, X2 = no)* | T1 ► T2 ► T5 ► T6 ► T7 ► T13 ► T14 | 255 | 325 (335) | 290 (295) |
| C "Surgical management"*(X1 = yes, X2 = yes)* | T1 ► T2 ► T5 ► T6 ► T8 ► ( T9 ∥ T10 ) ► (wait to satisfy TI2 & TI5) ► T11 ► T12 ► T13 ► T14 | 465 | 760 (770) | 603 (618) |

## 6 Conclusions

We find that an LLM like ChatGPT model o3 can serve as a smart assistant and conversational partner in the domain of business process analysis for models of reasonable complexity with nested gateway structures and inter-task constraints. The LLM acts just like a smart human assistant or analyst that can learn from feedback and correct mistakes. Even a novice user can have a rich conversational interaction with it. This shows that process design, analysis, and optimization are no longer the province of expert users. In the evaluation, we found that o3 is far superior to Gemini 2.5, Claude Opus 4.1 and Grok 3.

Overall, o3 is strong at reasoning clearly and could acquire a deep understanding of the processes at a semantic level beyond just understanding the process structure and relationships for processes of reasonable size and complexity from finance and healthcare domains. It can understand the syntax, semantics, and underlying logic of the business process at a deeper level. It can find errors in the process model, make corrections to fix the model, and make various calculations involving redesign scenarios. A user can provide an existing model to begin or even provide a text description from which the LLM can generate a process model.

Future work should undertake more robustness testing on still larger models. It would also be helpful to develop benchmarks as in [10] for testing and explore better user interfaces to assist novice users further.